# Environmental Sounds Spectrogram Classification using Log-Gabor Filters and Multiclass Support Vector Machines

Sameh Souli [1], Zied Lachiri [2]

[1]*Signal, Image and pattern recognition research unit*
*Dept. of Genie Electrique, ENIT*
*BP 37, 1002, Le Belvédère, Tunisia*

[2]*Dept. of Physique and Instrumentation, INSAT*
*BP 676, 1080, Centre Urbain, Tunisia*

**Abstract**

This paper presents novel approaches for efficient feature extraction using environmental sound magnitude spectrogram. We propose approach based on the visual domain. This approach included three methods. The first method is based on extraction for each spectrogram a single log-Gabor filter followed by mutual information procedure. In the second method, the spectrogram is passed by the same steps of the first method but with an averaged bank of 12 log-Gabor filter. The third method consists of spectrogram segmentation into three patches, and after that for each spectrogram patch we applied the second method. The classification results prove that the second method is the most efficient in our environmental sound classification system. These methods were tested on a large database containing 10 environmental sound classes. The best performance was obtained by using the multiclass support vector machines (SVM's), producing an average classification accuracy of 89.62 %.

## 1. Introduction

The research of environmental sound classification is less developed than that of speech and music classification. Recently, some efforts have been interested on classifying environmental sounds [4], [5] which the objective is to offer many services, for instance surveillance and security applications. In addition, the sound recognition systems used are based on different descriptors such as classic acoustic descriptors, cepstral descriptors, spectral descriptors, and time-frequency descriptors. These descriptors can be used as a combination of some, or even all, of these 1-D audio features together, but sometimes the combination between descriptors increases the classification performance compared with the individual used features.

Recently, some efforts emerge in the new research direction, which demonstrate that the visual techniques can be applied in musical [17].

In order to explore the visual information of environmental sounds, our last work consists of integrate the audio texture concept as image textures [18]. Our goal has to develop an environmental sounds classification method, using advanced visual descriptors. The feature extraction method uses the structure time-frequency by means of translation-invariant wavelet decomposition and a patch transform alternated with two operations: local maximum, global maximum to reach scale and translation invariance. In order to enhance this work, we develop here a nonlinear feature extraction method in the visual domain using in this case log-Gabor filters applied to spectrograms.

Besides, many studies likes [6], [19] show that spectro-temporal modulations play an important role in sound perception, and stress recognition in speech [20], in particular the 2D Gabor, which are suitable and very efficient to feature extraction.

In the recognition patterns, especially in image classification, Gabor filters are considerate as an efficient technique for obtaining a good feature. They offer an excellent simultaneous localization of spatial and frequency information [21]. They have many useful and important properties, in particular the capacity to decompose an image into its underlying dominant spectro-temporal components. The Gabor filters represent the most effective means of packing the information space with a minimum of spread and hence a minimum of overlap between neighboring units in both space and frequency [22].

In this paper we develop three new methods, based on spectro-temporal components. The First method begin by spectrogram calculation, which then was passed through a single log-Gabor filter, and finally passed through an optimal feature procedure based on mutual information. The second method is similar than the first method but in this case, with an averaged 12 log-Gabor filters. In the third method, we divide spectrogram into 3 patches, and then we apply second method for each spectrogram. In classification step, we use the SVM's with multiclass approach: One-Against-One.

This paper is organized as follows. Section 2 describes the background review of environmental sound classification system. Section 3 devotes environmental sound classification system using log-Gabor filters. Classification results are given in Section 4. Finally conclusions are presented in Section 5.

## 2. Background Review

Visual Feature extraction method for audio signal processing is reviewed in Sec. 2.1.
The visual domain was interested by speech researchers, especially by Victor Zue and his students [7] which demonstrate that the spectrogram was used to analyse the phoneme structure. Recently, some studies were adopted the visual methods in the musical sounds domain [17], [23], based on a technique inspired by image texture approach [8]. The proposed approach by [17] shows that the use of visual features for musical sounds obtains a good result for classification system. Of this fact, we had the idea to apply the visual features to environmental sounds. Indeed, the use of visual features makes the representation sparse, physically interpretable and the classification result very satisfactory. The advantages of this representation are the ability to capture the inherent structure within each type of environmental sound and to capture characteristics in the signal [4].
The feature method consists of four steps. First, a grey-scale spectrogram is generated from environmental sound which, passed in the translation-invariant wavelet transform phase (S1), to construct wavelet coefficients for three scales and three orientations. Then, we applied a local maximum (C1) for the obtained wavelet coefficients. After that we introduce a patch transform (S2), to group together the similar time-frequency geometries. Intuitively, for each patch, a global maximum (C2) is calculated, to select a representative time-frequency structure and to form feature vector for classification. This feature extraction method uses scale and translation invariance [8]. The overall classification system is shown in Fig. 1.
We illustrated the visual descriptors extraction step below [17].

- *Translation-invariant wavelet transform*

Let $s(x,y)$ be a spectrogram of the size $N_1 \times N_2$. We used the translation-invariant wavelet transform. The resulting wavelet coefficients will be defined by:

$$Wf(u,v,j,k) = \sum_{x=1}^{N_1} \sum_{y=1}^{N_2} s(x,y) \frac{1}{2^j} \varphi^k\left(\frac{x-u,y-v}{2^j}\right) \quad (1)$$

Where $k = 1,2,3$ is the orientation (horizontal, vertical, diagonal), $\varphi^k(x,y)$ is the wavelet function.

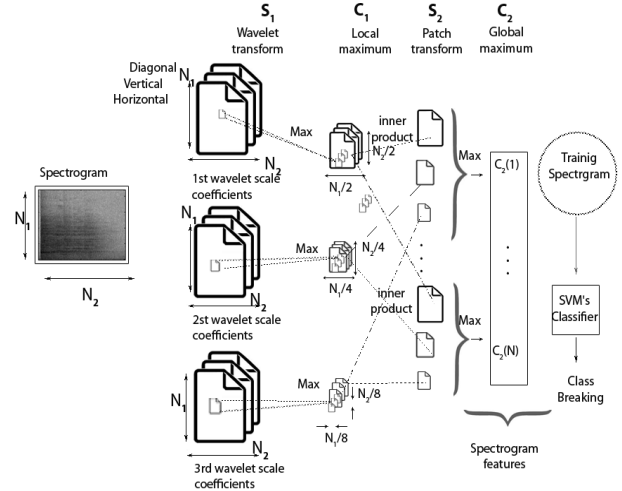

Fig. 1 Classification System Overview.

Fig. 2 shows the spectrogram of signal "dog bark" and the translation-invariant wavelet coefficients according the three spatial orientations: horizontal, vertical and diagonal for three scales.

Indeed, to build a translation-invariant wavelet representation, the scale is made discrete but not the translation parameter. The scale is sampled on a dyadic analysis $\{2^j\}_{j \in Z}$. The use of the translation-invariant wavelet transform creates a redundancy of information that allows keeping the translation-invariance at all levels of factorization [1].
The scale invariance is carried out by normalization, using the following formula:

$$S_1(u,v,j,k) = \frac{|Wf(u,v,j,k)|}{\|S\|^2_{supp(\varphi_j^k)}} \quad (2)$$

Where $\|S\|^2_{supp(\varphi_j^k)}$ is the energy of spectrogram detail wavelet coefficients.
The scale invariance is the special feature of keeping the same appearance when we "zoom" the images whatever the scale at which they are observed.
In fact, the wavelet analysis or the multiresolution analysis are good tools for the analysis of scaling laws, thus helping to emphasize and characterize a scale invariance in a reliable way [1]. The introduction of the properties of scale invariance then leads to new multi-resolution spaces.

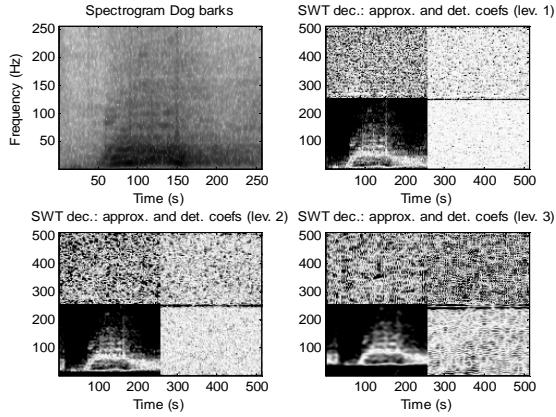

Fig. 2 Representation of the Translation-invariant wavelet coefficients for three Orientations and three Levels of scales.

- Local Maximum

The continuation of translation invariance [8] is done by calculating the local maximum of $S_1$ :

$$C_1(u,v,j,k) = \max_{u' \in [2^j(u-1)+1, 2^j u], v' \in [2^j(v-1)+1, 2^j v]} S_1(u',v',j,k) \quad (3)$$

The $C_1$ section is obtained by a subsampling of $S_1$ using a cell grid of the $2^j \times 2^j$ size that is then followed by the local maximum. Generally, the maximum being taken at each $j$ scale and $k$ direction of a spatial neighborhood of a size that is proportional to $2^j \times 2^j$. The resulting $C_1$ at the $j$ scale and the $k$ direction is therefore of the $N_1/2^j \times N_2/2^j$ size, where $j = 1,2,3$.

- Patch Transform

Mallat and Peyré [9] proposed in their researches the grouplet transform by using the Haar transform on the wavelet coefficients, which consists in replacing two neighbors' coefficients $(a,b)$ by their mean and their difference. Inspired by this method, the idea consists of selecting $N$ patch $P_i$, then the scalar product is calculated between these patch $P_i$, and the $C_1$ coefficients, followed by a sum. Indeed, for every patch, we get only one scalar:

$$S_2(u,v,j,i) = \sum_{u'=1}^{N_1/2^j} \sum_{v'=1}^{N_2/2^j} \sum_{k=1}^{3} C_1(u',v',j,k) P_i(u'-u, v'-v, k)$$
(4)

Where $P_i$ of size $M_i \times M_i \times 3$ are the patch functions that group 3 wavelet orientations. The patch functions are extracted by a simple sampling at a random scale and a random position of the $C_1$ Coefficients of a spectrogram [8], for instance a $P_0$ patch of the $M_0 \times M_0$ size contains $M_0 \times M_0 \times 3$ elements, $M_0$ may take the following values ($M_0 = 4,8,12$).

- Global maximum

The $C_2$ coefficients are obtained by the application of the max function on $S_2$:

$$C_2(i) = \max_{u,v,j} S_2(u,v,j,i) \quad (5)$$

In this work, the obtained result is a vector of $NC_2$ values, where $N$ corresponds to the number of extracted patches. In this way, the $C_2$ obtained coefficients constitute the parameter vector for the classification with SVM.

## 3. Environmental sound classification system with Log-Gabor Filters

Our environmental sound classification system consists of three methods. In the first method, a spectrogram is generated from sound [10]. Next, it passed to single log-Gabor filter extraction. Then, we applied mutual information in order to get an optimal feature. This feature is finally used in the classification.

The second method consists of the same steps as first method, but with an averaged 12 log-Gabor filters instead of single log-Gabor filter.

In the third method the idea is to segment each spectrogram into 3 patches. Intuitively, for each patch, an averaged 12 log-Gabor filters are calculated, after that we applied a mutual information selection to pass then in the classifier. In classification phase, we use SVM, in One-Against-One configuration with the Gaussian kernel.

2.1. Feature extraction methods

The feature extraction is based on three methods. These methods use the log-Gabor filters.

2.1.1. Single log-Gabor filter

The procedure for generating the single log-Gabor filter is shown in Fig. 3.This approach consists in computation of 12 log-Gabor filters that are derived from the environmental sounds spectrograms, with 2 scales (1,2) and 6 orientations (1,2,3,4,5,6), this extraction allows the best correlate of signal structures. Then, for each single filter result we calculated the magnitude, after that, we passed through on mutual information (MI) algorithm to find an optimal feature vector (Fig.3) that next passed for classification phase [20].

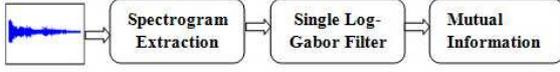

Fig. 3 Feature extraction using single log-Gabor filter.

2.1.2. 12 log-Gabor Filters concatenation

In this method, each environmental sound spectrogram was passed thought a bank of 12 log-Gabor filters. This produced a bank of 12 log-Gabor filters $\{G_{11}, G_{12}, \ldots, G_{16}, G_{21}, \ldots, G_{25}, G_{26}\}$, with each filter representing different scale and orientation. Thus, this result allows us to say that we obtain for each spectrogram a bank of 12 log-Gabor filters. These resulting feature values were next concatenated into 1D-vectors. Then the averaged computation, passed thought the MI criteria, and was sent to SVM for classification (Fig. 4).

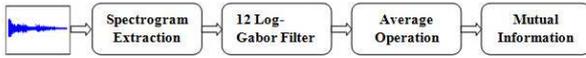

Fig. 4 Feature extraction using 12 log-Gabor filters.

2.1.3. Three Spectrogram Patches with 12 log-Gabor Filters

The method concept is to use the spectrogram patch, the aim is to find the suitable part of spectrogram, where concentrates the efficient structure, which gives a better result. The idea is to extract three patches from each spectrogram. The first patch included frequencies from 0.01Hz to 128Hz, the second patch, from 128Hz to 256Hz, and the third patch, from 256Hz to 512Hz. Indeed, each patch was passed through 12 log-Gabor filters, followed by an averaged operation and then passed to MI feature selection algorithm, which constitute the parameter vector for the classification (Fig. 5).

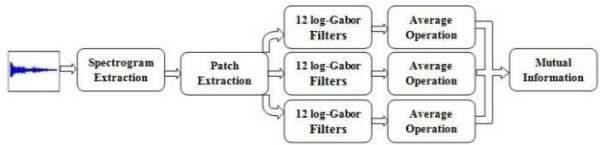

Fig. 5 Feature extraction using 3 spectrogram patches with12 log-Gabor filters.

2.2. Environmental Sound Spectrogram

The spectrogram is the most current time-frequency representation. It is a visual energy representation across frequencies and over time. The horizontal axis represents time, and the vertical axis is frequency [11].

With spectrogram we can observe the complete spectrum of environmental sounds and express sound by combining the merit of time and frequency domains [24]. Furthermore, we can easily identify the environmental sounds spectrograms by their contrast, since they are considered as different textures Fig. 6 [23].These observations show that the spectrograms contain characteristics which can be used to differentiate between different environmental sounds class [21].

The sound time-frequency contains a large amount of information and provides a representation that can be easily interpreted [7].The Short-Time Fourier Transform (STFT) was used to calculate the spectrogram $s(x, y)$, and the frames were taken to be 256-point frames with 192-point overlap.

Let $f[n]$ be an audio signal, $n = 0,1, \ldots, N − 1$.

The time-frequency transform factorizes f over a family of time-frequency atoms $\{g_{x,y}\}_{x,y}$ where $x$ and $y$ are respectively time and frequency. The short-time Fourier transform of f is defined by [10]:

$$F[x, y] = \langle f, g_{x,y} \rangle = \sum_{n=0}^{N-1} f[n]\, g_{x,y}^*[n] \qquad (6)$$

where $*$ is the conjugate. The atoms of the short-time Fourier transform are:

$$g_{x,y}[n] = w[n − l]\exp\left(\frac{i2\pi kn}{k}\right) \qquad (7)$$

where $w[n]$ is the Hamming window, for each
$0 \le y < k$, $F[x, y]$ is calculated for $0 \le y < k$. The classification is based on the log-spectrogram:

$$s(x, y) = \log|F[x, y]| \qquad (8)$$

Let us take the spectrograms of environmental sounds as illustrated in Fig. 6, each class contains sounds with very different temporal or spectral characteristics, levels, duration, and time alignment for example door slams present a wide frequency band but with a short duration.

In addition, for the children voices we can distinguish the presence of the privilege frequencies. Concerning phone rings, we remark that it presents fundamental frequencies. Another remark about phone rings and children voices, they are harmonic sounds. Furthermore, we notice that there are some similarities between explosions and gunshots though, they belong to different classes.

We also illustrate according to Fig. 6 that there are signals which present textural properties can be easily learned without explicit detailed analysis of the corresponding patterns [5], so easy to be distinguished, which influences in a positive way in the phase of the classification.

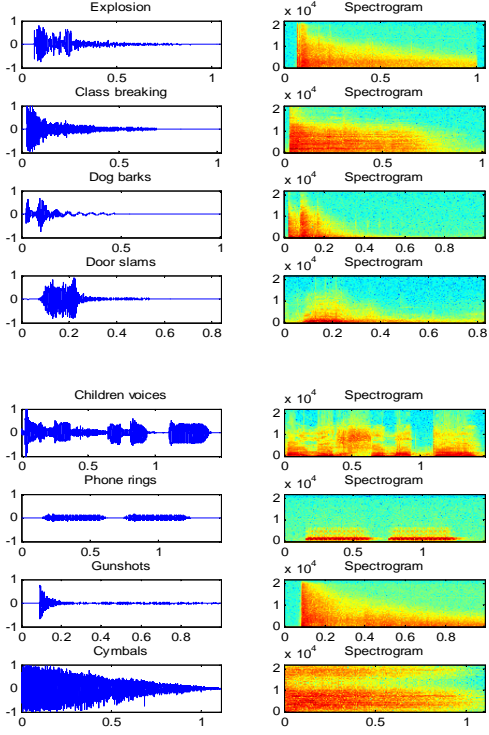

Fig. 6 Audio waveform and Spectrograms of 8 classes environmental sound.

### 2.3. Log-Gabor-filters

Gabor filters offer an excellent simultaneous localization of spatial and frequency information [21]. They have many useful and important properties, in particular the capacity to decompose an image into its underlying dominant spectro-temporal components [6]. The log-Gabor function in the frequency domain can be described by the transfer function $G(r, \theta)$ with polar coordinates [20]:

$$G(r, \theta) = G_{radial}(r) \cdot G_{angular}(r) \qquad (9)$$

Where $G_{radial}(r) = e^{-\log(r/f_0)^2 / 2\sigma_r^2}$, is the frequency response of the radial component and $G_{angular}(r) = exp\left(-(\theta/\theta_0)^2 / 2\sigma_\theta^2\right)$, represents the frequency response of the angular filter component.
We note that $(r, \theta)$ are the polar coordinates, $f_0$ represents the central filter frequency, $\theta_0$ is the orientation angle, $\sigma_r$ and $\sigma_\theta$ represent the scale bandwidth and angular bandwidth respectively.

The log-Gabor feature representation $|S(x, y)|_{m,n}$ of a magnitude spectrogram $s(x, y)$ was calculated as a convolution operation performed separately for the real and imaginary part of the log-Gabor filters:

$$Re(S(x,y))_{m,n} = s(x,y) * Re(G(r_m, \theta_n)) \qquad (10)$$
$$Im(S(x,y))_{m,n} = s(x,y) * Im(G(r_m, \theta_n)) \qquad (11)$$

$(x, y)$ represent the time and frequency coordinates of a spectrogram, and $m = 1, \dots, N_r = 2$ and $n = 1, \dots, N_\theta = 6$ where $N_r$ devotes the scale number and $N_\theta$ the orientation number. This was followed by the magnitude calculation for the filter bank outputs:

$$|S(x,y)| = \sqrt{\left(Re(S(x,y))_{m,n}\right)^2 + Im(S(x,y))_{m,n}} \qquad (12)$$

### 2.4. Averaging outputs of log-Gabor filters.

The averaged operation was calculated for each 12 log-Gabor filter, appropriate for each spectrogram, which purpose is to obtain a single output array [20]:

$$|\hat{S}(x,y)| = \frac{1}{N_r N_\theta} \sum_{\substack{m=1 \\ n=1}}^{N_r, N_\theta} |S(x,y)|_{m,n} \qquad (13)$$

### 2.5. Features optimization using mutual information.

The information found commonly in two random variables is defined as the mutual information between two variables X and Y, and it is given as [12]:

$$I(X;Y) = \sum_{x \in X} \sum_{y \in Y} p(x,y) \log \frac{p(x,y)}{p(x)p(y)} \qquad (14)$$

Where $p(x) = Pr(X = x)$ is the marginal probability density function and $p(x) = Pr(X = x)$, and $p(x,y) = Pr(X = x, Y = y)$ is the joint probability density function.

### 2.6. SVM Classification

For the classification, we employ a Support Vector Machines, in a One-against-One configuration [13].
Let a set of data $(x_1, y_1), \dots, (x_m, y_m) \in \Re^d \times \{\pm 1\} \in$ where $X = \{x_1, \dots, x_m\}$ a dataset in $\Re^d$ where each $x_i$ is the feature vector of a signal. In the nonlinear case, the idea is to use a kernel function $K(x_i, x_j)$, where $K(x_i, x_j)$ satisfies the Mercer conditions [14]. Here, we used a Gaussian RBF kernel witch formula is:

$$k(x, x') = exp\left[\frac{-\|x - x'\|^2}{2\sigma^2}\right]. \qquad (15)$$

Where $\|.\|$ indicates the Euclidean norm in $\Re^d$.

Let $\Omega$ be a nonlinear function which transforms the space of entry $\Re^d$ to an intern space $H$ called a feature space. $\Omega$ allows to perform a mapping to a large space in which the linear separation of data is possible [2].

$$\Omega: \Re^d \longrightarrow H$$
$$(x_i, x_j) \mapsto \Omega(x_i)\Omega(x_j) = k(x_i, x_j) \quad . \quad (16)$$

The $H$ space is a reproducing kernel Hilbert space (RKHS) of functions.
Thus, the dual problem is presented by a Lagrangian formulation as follows:

$$\max W(\alpha) = \sum_{i=0}^{m} \alpha_i - \frac{1}{2}\sum_{i,j=1}^{m} y_i y_j \alpha_i \alpha_j k(x_i, x_j)|_{i=1,\dots,m} \quad (17)$$

Under the following constraints:

$$\sum_{i=1}^{m} \alpha_i y_i = 0, \quad 0 \leq \alpha_i \leq C. \quad (17)$$

They $\alpha_i$ are called Lagrange multipliers and $c$ is a regularization parameter which is used to allow classification errors. The decision function will be formulated as follows:

$$f(x) = sgn(\sum_{i=1}^{m} \alpha_i y_i k(x, x_i) + b) \quad (18)$$

We hence adopted one approach of multiclass classification: One-against-One. This approach consists of creating a binary classification of each possible combination of classes, and the result for $k$ classes is $k(k-1)/2$. The classification is then carried out in accordance with the majority voting scheme [16].

## 4. Experimental Evaluation

### 4.1. Experimental Setup

Our corpus of sounds comes from commercial CDs [26]. Among the sounds of the corpus we find: explosions, broken glass, door slamming, gunshot, etc.
This database includes impulsive and harmonic sounds. We used 10 classes of environmental sounds as shown in Table 1.

All signals have a resolution of 16 bits and a sampling frequency of 44100 Hz that is characterized by a good temporal resolution and a wide frequency band.
Most of the signals are impulsive; we took 2/3 for the training and 1/3 for the test.

Among the big problems met during the classification by the SVM's is the choice of the values of the kernel parameter $\gamma$ and the constant of regularization $C$. To resolve this problem we suggested the cross-validation procedure [3]. Indeed, according to [25], this method consists in setting up a grid-search for $\gamma$ and C. For the implementation of this grid, it is necessary to proceed iteratively, by creating a couple of values $\gamma$ and C.
The radial basis kernel was adopted for all the experiments. The parameter C was used also for determined the tradeoff between margin maximization and training error minimization [15].

Table 1: Classes of sounds and number of samples in the database used for performance evaluation.

| Classes | Train | Test | Total |
|---|---|---|---|
| Door slams (Ds) | 208 | 104 | 312 |
| Explosions (Ep) | 38 | 18 | 56 |
| Class breaking (Cb) | 38 | 18 | 56 |
| Dog barks (Db) | 32 | 16 | 48 |
| Phone rings (Pr) | 32 | 16 | 48 |
| Children voices (Cv) | 54 | 26 | 80 |
| Gunshots (Gs) | 150 | 74 | 224 |
| Human screams (Hs) | 48 | 24 | 72 |
| Machines (Mc) | 38 | 18 | 56 |
| Cymbals (Cy) | 32 | 16 | 48 |
| Total | 670 | 330 | 1000 |

### 4.2 Experimental Results

The results of the first method are summarized in Table 2, the classification rates for each single log-Gabor filter, which included 2 scales and 6 orientations, are relatively low, ranging from 42.85% to 99.67% for 10 sounds class.
The best classification result based on first method belongs to the Door slams class with scale=1, and orientation=3.
To improve the first method result, features should be extracted either from all log-Gabor filters or from a selected group of best performing filters [20]. Both the second and the third method are concentrated to show them.
Results of the second approach and third approach are illustrated in Table 3.

Table 2: Recognition Rates of 12 log-Gabor filters applied to one-against-one SVM's based classifier with Gaussian RBF kernel

| Scale | Orientation | Ds | Ep | Cb | Db | Pr | Cv | Gs | Hs | Mc | Cy |
|---|---|---|---|---|---|---|---|---|---|---|---|
| 1 | 1 | 99,35 | 46.42 | 57.14 | 83.33 | 68.75 | 82.50 | 89.28 | 88.23 | 80.35 | 93.75 |
|   | 2 | 96.15 | 48.21 | 60.71 | 79.16 | 70.83 | 77.50 | 89.73 | 89.70 | 82.14 | 89.58 |
|   | 3 | 99.67 | 42.85 | 66.07 | 77.08 | 72.91 | 80.00 | 91.07 | 91.17 | 83.92 | 87.50 |
|   | 4 | 99.03 | 44.64 | 67.85 | 81.25 | 77.08 | 78.75 | 88.39 | 92.64 | 78.57 | 85.41 |
|   | 5 | 98.39 | 46.42 | 55.35 | 79.16 | 66.66 | 72.50 | 86.16 | 86.76 | 76.78 | 83.33 |
|   | 6 | 99.03 | 42.85 | 51.78 | 81.25 | 64.58 | 71.25 | 85.71 | 73.52 | 75.00 | 81.25 |
| 2 | 1 | 99,35 | 62.50 | 71.42 | 87.50 | 83.33 | 85.00 | 95.98 | 89.70 | 89.28 | 95.83 |
|   | 2 | 99.35 | 64.28 | 75.00 | 89.58 | 85.41 | 87.50 | 96.42 | 92.64 | 87.50 | 85.41 |
|   | 3 | 80.12 | 64.28 | 78.57 | 91.66 | 87.50 | 86.25 | 95.08 | 94.11 | 85.71 | 87.50 |
|   | 4 | 83.01 | 44.64 | 75.00 | 79.16 | 81.25 | 82.50 | 94.64 | 85.29 | 82.14 | 83.33 |
|   | 5 | 80.44 | 55.35 | 67.85 | 77.08 | 79.16 | 81.25 | 90.62 | 80.88 | 83.92 | 81.25 |
|   | 6 | 81.08 | 58.92 | 66.07 | 77.08 | 75.00 | 77.50 | 89.73 | 76.47 | 69.64 | 81.25 |

Table 3: Recognition Rates for averaged outputs of 12 log-Gabor filters, 3 Spectrogram patches and descriptors with wavelet-transform applied to one-against-one SVM's based classifier with Gaussian RBF kernel

| Classes | 12 log-Gabor filters | 3 Spectrogram patches | descriptors with wavelet-transform |
|---|---|---|---|
| Ds | 99,35 | 94.87 | 94.28 |
| Ep | 62.50 | 69.64 | 94.28 |
| Cb | 78.57 | 78.57 | 97.43 |
| Db | 87.50 | 89.58 | 88.88 |
| Pr | 83.33 | 87.50 | 83.33 |
| Cv | 87.50 | 82.50 | 93.33 |
| Gs | 98.21 | 83.03 | 97.61 |
| Hs | 94.11 | 95.58 | 92.59 |
| Mc | 89.28 | 92.85 | 90.47 |
| Cy | 95.83 | 93.75 | 88.88 |

Indeed, let us begin by the second method, which the idea consists of 12 log-Gabor filters concatenation, and then an averaged operation is applied, followed by the mutual information criteria. The obtained classification results are better than the classification results attained by a single log-Gabor filter method and range from 62.50% to 99.35%. We were able to achieve an averaged accuracy rate of the order 89.62% in ten classes with one-against-one approach. In the third approach results, we obtained an averaged accuracy rate of the order 86.78%.
This result is better than the first method result, but is slightly lower than the second method result.
The experiments results are satisfactory, so this fact encourage us to investigate better in the visual domain.

### 4.3 Comparison of Visual Descriptors

We compare the overall recognition accuracy using 12 log-Gabor-filters concatenation method, three spectrogram patches with 12 log-Gabor filters method and visual descriptors with wavelet-transform in Table 3. As shown in this table, 12 log-Gabor filters features possess the best recognition rate which belongs in the Door slams class. This method perform better than three spectrogram patches with 12 log-Gabor filters in five of the examined classes while producing poor results in the case of four other classes.
In the other case, the comparison between visual descriptors with wavelet-transform and 12 log-Gabor filters features method shows that the last method is very high, in five classes but is slightly low in other five classes.
The 12 log-Gabor filters features perform better overall, with the exception of two classes (Explosions (Ep), Class breaking (Cb)) having the lowest recognition rate at 62.50%. With 12 log-Gabor filters feature, we were able to achieve an averaged accuracy rate of 89.62% in discriminating ten classes. There are four classes that have a classification rate higher than 90%. Concerning visual descriptors with wavelet-transform, we attained an averaged accuracy rate of 91.82% in the same discriminating ten classes.
We see that 12 log-Gabor filters feature and visual descriptors with wavelet-transform obtain a good performance in the visual domain.
We can conclude that using descriptors belongs to visual domain provides us with extra information for discriminating between difficult classes.

## 5. Conclusion

In this paper, we propose three new methods for environmental sound classification, based on visual domain. We show how these methods are efficient to classify the environmental sounds. All methods use log-Gabor filters, but with 3 different manners. The first method uses a single log-Gabor filter. The second method uses an averaged 12 log-Gabor filters concatenation. The third method segmented spectrogram into three patches with averaged 12 log-Gabor filters. The important point of these methods is to present an improved feature set including visual features.
We prove that the second method obtain the best averaged classification result of the order 89.62%. The obtained results are very satisfactory in the visual domain.

These results need more exploration. The proposed approaches can be improved while digging deeply into the visual domain. Future research directions will include another methods extracted from image processing.

**Acknowledgments**

We are grateful to G. Yu for many discussions by mail.